\documentclass[conference]{IEEEtran}
\IEEEoverridecommandlockouts

\usepackage{cite}
\usepackage{amsmath,amssymb,amsfonts}
\usepackage{algorithmic}
\usepackage{graphicx}
\usepackage{textcomp}
\usepackage[table]{xcolor}
\usepackage{xspace}
\usepackage{booktabs}
\usepackage{array}
\usepackage{multirow}

\usepackage{bm}
\usepackage{float}
\usepackage{wrapfig}
\usepackage{color}
\usepackage{amssymb}
\usepackage{bbding}
\usepackage{diagbox}
\usepackage{colortbl}
\usepackage{enumitem}

\newcommand{\PreserveBackslash}[1]{\let\temp=\\#1\let\\=\temp}
\newcolumntype{C}[1]{>{\PreserveBackslash\centering}p{#1}}
\newcolumntype{R}[1]{>{\PreserveBackslash\raggedleft}p{#1}}
\newcolumntype{L}[1]{>{\PreserveBackslash\raggedright}p{#1}}

\def\BibTeX{{\rm B\kern-.05em{\sc i\kern-.025em b}\kern-.08em
    T\kern-.1667em\lower.7ex\hbox{E}\kern-.125emX}}

\newcommand{\method}{MTEEG\xspace}

\usepackage{amsmath,amsfonts,bm}

\def\eqref#1{equation~\ref{#1}}

\def\1{\bm{1}}

\def\vx{{\bm{x}}}

\def\mW{{\bm{W}}}

\DeclareMathAlphabet{\mathsfit}{\encodingdefault}{\sfdefault}{m}{sl}
\SetMathAlphabet{\mathsfit}{bold}{\encodingdefault}{\sfdefault}{bx}{n}

\newcommand{\R}{\mathbb{R}}

\begin{document}

\title{Towards Unified Multi-task EEG Analysis with Low-Rank Adaptation\\
}

\author{\IEEEauthorblockN{Sicheng Dai\textsuperscript{1,2,3,4},
Kai Chen\textsuperscript{4},
Hongwang Xiao\textsuperscript{4,5}, 
Shan Yu\textsuperscript{1,3}, and
Qiwei Ye\textsuperscript{4}}
\IEEEauthorblockA{\textsuperscript{1}Institute of Automation, Chinese Academy of Sciences}
\IEEEauthorblockA{\textsuperscript{2}School of Artifcial Intelligence, University of Chinese Academy of Sciences}
\IEEEauthorblockA{\textsuperscript{3}State Key Laboratory of Brain Cognition and Brain-inspired Intelligence Technology}
\IEEEauthorblockA{\textsuperscript{4}Beijing Academy of Artificial Intelligence}
\IEEEauthorblockA{\textsuperscript{5}State Key Laboratory of Multimedia Information Processing, Peking University}
\IEEEauthorblockA{daisicheng2023@ia.ac.cn, chenkai.cn@hotmail.com, hwxiao@baai.ac.cn, shan.yu@nlpr.ia.ac.cn, qwye@baai.ac.cn}%

}

\maketitle

\begin{abstract}
Recent self-supervised pre-training methods for electroencephalogram (EEG) have shown promising results. However, the pre-trained models typically require full fine-tuning on each downstream task individually to achieve good performance. In practical applications involving multiple tasks, utilizing a separate model for each task is not ideal regarding computational and spatial cost. In this study, we go one step further and explore the simultaneous adaptation of a pre-trained model to multiple different tasks. The EEG signals exhibit significant heterogeneity due to their collection from various subjects using diverse devices and experimental setups, resulting in potential conflicts among different tasks that impede joint optimization. To tackle this challenge, we propose \method, a multi-task EEG analysis framework which incorporates task-specific low-rank adaptation (LoRA) modules to disentangle the parameter space and alleviate task conflicts. To investigate the trade-off between task specification and interaction, we propose three variants of \method that integrate the LoRA modules in different ways and evaluate them on six downstream tasks, demonstrating that \method can surpass state-of-the-art single-task methods on the majority of metrics. \method shows the potential of multi-task EEG analysis and promotes the development of general-purpose brain-computer interfaces in the future.
\end{abstract}

\begin{IEEEkeywords}
EEG, brain-computer interface, multi-task learning
\end{IEEEkeywords}

\section{Introduction}

Electroencephalography (EEG) is a widely used neuroimaging technique that captures electrical activity of the brain through non-invasive scalp electrodes. In recent years, deep learning models, such as convolutional neural networks (CNNs) and transformers, have demonstrated remarkable success in extracting meaningful patterns from EEG data, leading to significant improvements in various applications including emotion recognition, motor imagery classification \cite{li2022eeg} and seizure detection \cite{boonyakitanont2020review}. However, despite their capability, these models are typically customized for specific tasks and input formats, which causes them to overfit and become ungeneralizable.

\begin{figure}[t]
  \centering
  \hspace*{-0.2cm}
  \includegraphics[width=0.49\textwidth]{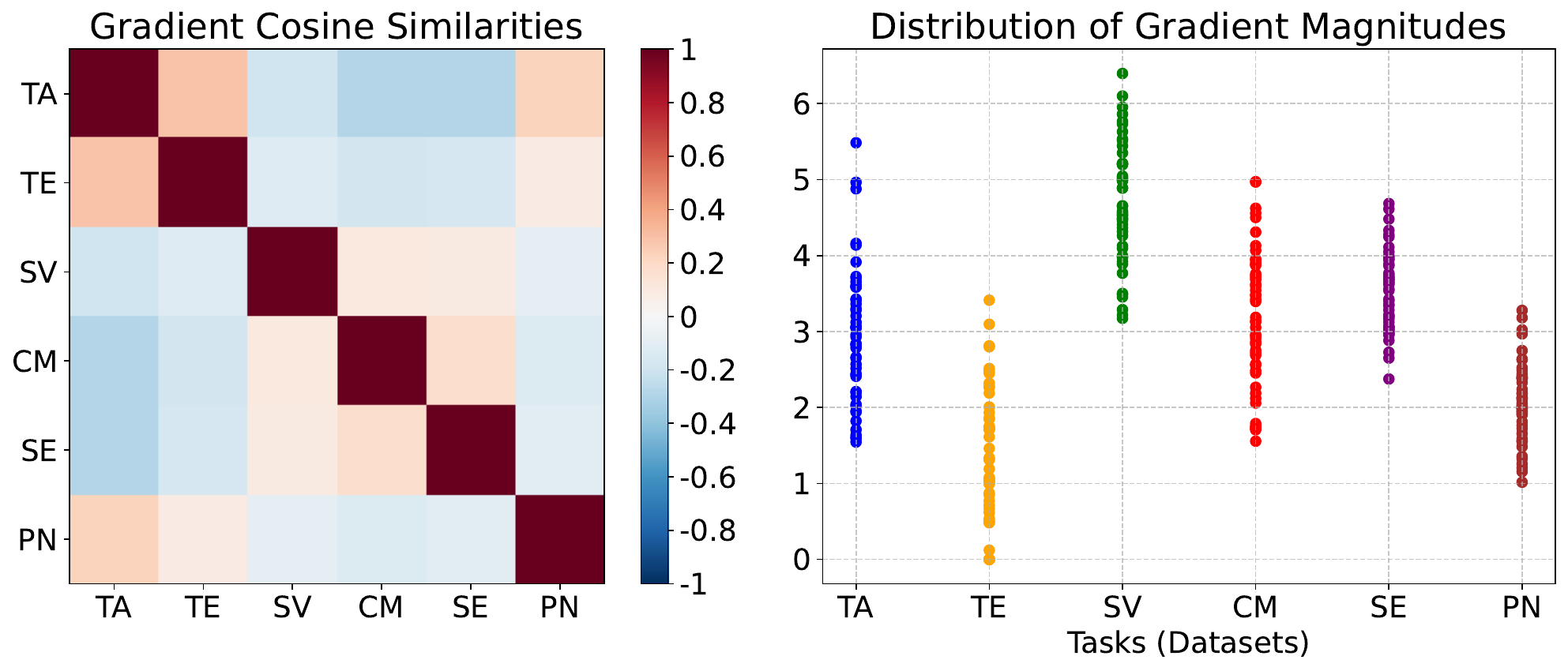} 
      \caption{Visualization of gradients from each task in a hard parameter sharing (HPS) framework with LaBraM backbone. (Left) Gradient cosine similarities between each two of the six tasks. (Right) Distribution of the gradient magnitudes from each task. The x-axis is labeled with the abbreviated names of datasets.}
    \label{fig:grad_sim}
    \vspace{-0.5em}
\end{figure}

Drawing inspirations from the advancements of large language models \cite{devlin2018bert,achiam2023gpt}, some researchers \cite{yang2023biot,yi2024learning,jiang2024large} employ self-supervised learning to extract generic representations from large amounts of unlabeled EEG data, significantly improving the model's generalizability. Despite their remarkable performance, these models necessitate individual fine-tuning for each downstream task, thereby constraining their versatility and applicability in practical scenarios involving multiple tasks. For example, an EEG-based health monitoring system may need to simultaneously perform multiple tasks, including seizure detection, emotion recognition and sleep stage classification, to have a comprehensive evaluation of patients' condition. In this case, a pre-trained model must be replicated and fine-tuned three times, once for each task, resulting in significant computational and spatial overhead. Therefore, it would be beneficial to have a unified system that is capable of handling different tasks concurrently.

Despite the promise, challenges persist to build an efﬁcient multi-task model for EEG processing. The EEG signals, collected from various subjects utilizing different devices and experimental configurations, exhibit markedly distinct intrinsic characteristics. This variability can mislead the model with conflicting parameter update directions (Figure \ref{fig:grad_sim}), leading to a substantial decrease in learning efficacy. Similar heterogeneity-induced issues have also been noted in other domains \cite{yu2020gradient,zhou2024exploring}, and many methods have been proposed to tackle them. For instance, some works incorporate separate modules for specific tasks \cite{liu2022polyhistor,mahabadi2021parameter}, while others use soft-gating mechanisms to flexibly assign modules for different tasks \cite{ma2018modeling,cheng2016cgc}. Nevertheless, the majority of these studies focus on the analysis of image, text and audio data, raising doubts about the applicability of their findings to EEG.

In this study, we propose \method, a novel EEG analysis framework which exploits a pre-trained LaBraM \cite{jiang2024large} along with task-specific low-rank adaptation (LoRA) modules to facilitate efficient multi-task joint training. To investigate the trade-off between task specification and interaction, we conduct experiments with three variants of \method that integrates the LoRA modules in different ways: 1) \method-SP allocates a separate LoRA module for each task, thereby maximizing task specification; 2) \method-RT employs a Mixture of Experts (MoE)-like design and reuses the same set of LoRA modules (experts) across all the tasks, with learnable routers to determine the weights of experts at each layer, thereby enhancing task interaction; 3) \method-DC decomposes the LoRA module into a task-agnostic down-projection matrix and multiple task-specific up-projection matrices, promoting the dual objectives of global knowledge reuse and task-specific knowledge disentanglement. Experiments show that \method-DC performs better than the other two variants and surpasses state-of-the-art single-task methods on the majority of tasks and metrics. Subsequent analysis reveals that \method-DC delineates the clearest task boundaries within its feature space, confirming its strong performance and capacity to alleviate task conflicts.
In summary, our contributions are as follows:\begin{itemize}
    \item We investigate multi-task EEG analysis, which is a crucial yet underexplored aspect in the practical application of brain-computer interfaces. Concurring with prior research on other data types, we observe that joint training on heterogeneous EEG datasets also presents the issue of conflicts between different tasks, leading to substantial performance deterioration of the model.
    \item We present the \method framework, which enhances a pre-trained model by incorporating task-specific modules to achieve parameter isolation across different tasks. This isolation allows for the separation of gradients to prevent conflicts, hence facilitating multi-task joint training. To take both task specification and interaction into account, we introduce three variants of the framework: \method-SP, \method-RT and \method-DC, and evaluate their performance on downstream tasks.
    \item Through extensive experiments, we demonstrate that after joint optimization on six publicly available datasets, \method can handle abnormal detection, event type classification, emotion recognition, seizure detection, sleep stage classification and motor imagery classification simultaneously, achieving performance superior than state-of-the-art single-task methods on the majority of metrics. 
\end{itemize}

\begin{figure*}[t]
    \centering
    \includegraphics[width=0.98\linewidth]{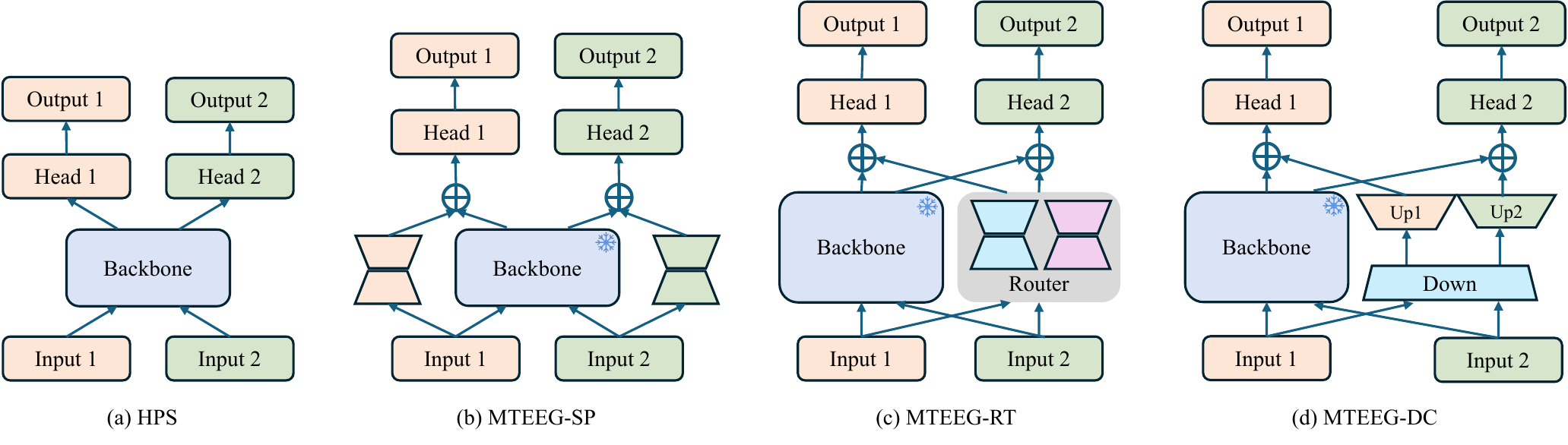}
    \vspace{-0.3cm}
    \caption{A comparison between hard parameter sharing (HPS) and the proposed framework. (a) HPS lets different tasks share the same modules except for the classification heads. (b) \method-SP isolates the parameters by allocating a separate LoRA module for each task. (c) \method-RT encourages task interaction by reusing the same set of experts for all tasks. (d) \method-DC balances task specification and interaction by combining a shared down-projection matrix and task-specific up-projection matrices. Note that in HPS all the modules are trainable, while in \method the pre-trained backbone network is kept frozen.}
    \vspace{-0.2cm}
    \label{fig:method}
\end{figure*}

\section{Methodology}

\subsection{Problem Formulation}
Assume there are a total of $P$ tasks. For $p \in \{1, 2, \dots, P\}$, given any multi-channel EEG signal $X \in \R^{C_p \times T_p}$ in the $p$-th task, where $C_p$ and $T_p$ represent the number of channels and the input duration respectively, the model aims to predict the corresponding label $y \in \mathcal{Y}_p$, where $\mathcal{Y}_p$ represents the set of all possible outputs.

\subsection{LaBraM Preliminaries}
\label{arch}
The architecture of \method is built upon that of LaBraM \cite{jiang2024large}. An input EEG sample $X \in \R^{C_p \times T_p}$ is first segmented in the temporal dimension with a non-overlapping window of length $w$, resulting in patches $\vx = \{x_{i,j}|i=1, 2, \dots, C_p, j=1, 2, \dots, \lfloor \frac{T_p}{w} \rfloor \}$. The patches are then processed sequentially by the temporal encoder, transformer encoder and classification head to produce the final output.

\subsubsection{Temporal Encoder}
The temporal encoder takes the segmented input patches and encode them into embeddings, serving to capture the intricate temporal features in the signal. It consists of multiple temporal convolution blocks, each of which is composed of a 1-D convolution layer, a group normalization layer, and a GELU activation function. Formally, given a set of input patches $\vx$, the output can be denoted as
\begin{align*}
    \{e_{i, j} = TE(x_{i, j}) \in \R^d|x_{i, j} \in \vx  \},
\end{align*}
where $TE$ represents the temporal encoder and $d$ is the dimension of the embeddings.

\subsubsection{Transformer Encoder}
To take account of the global features in the signal, the patch embeddings are added with temporal and spatial embeddings based on the 10-20 international system, then fed into the transformer encoder to be processed with the attention mechanism. The attention function can be formulated as
\begin{align*}
    {\rm Attention}(Q,K,V) = {\rm softmax}(\frac{{\rm LN}(Q){\rm LN}(K)^T}{\sqrt{d_p}})V,
\end{align*}
where $d_p$ is the dimension of the key and query, and LN stands for layer normalization, which are added to stabilize training by avoiding overly large values in the attention logits.

\subsubsection{Pre-training Procedure}
Before multi-task learning on downstream tasks, a LaBraM model is pre-trained on unlabeled data to provide a solid foundation for extracting useful information raw EEG signals. Specifically, we start by training a neural tokenizer which is inspired by VQ-VAE \cite{van2017neural}. The tokenizer is followed by a neural codebook which quantizes the continuous representations into discrete tokens. The learning process is then guided by the reconstruction of the amplitude and phase from these discrete tokens. After the tokenizer is sufficiently trained, we train the LaBraM model by randomly masking a proportion of the input patches and letting the model predict their corresponding indices in the codebook.

 \subsection{Multi-task Learning with LoRA}
 After pre-training, the model is adapted to downstream tasks via a fine-tuning process, in which LoRA modules are incorporated to achieve parameter isolation. To take account of the trade-off between task specification and interaction, we introduce three variants of \method that integrates the LoRA modules in different ways, as illustrated in Figure \ref{fig:method}.

 \begin{table}[th]
    \centering
    \caption{A summary of the downstream dataset statistics}
    \label{tab:downstream}
    \resizebox{0.48\textwidth}{!}{
    \setlength{\tabcolsep}{0.7mm}{
    \hspace*{-0.3cm}
    \begin{tabular}{C{1.8cm}C{1.8cm}C{1.8cm}C{1.8cm}C{1.8cm}}
    \toprule[1.5pt]
  \textbf{Dataset} & \textbf{\# Channel} & \textbf{Sampling Rate (Hz)}  & \textbf{Duration (seconds)} & \textbf{\# Sample} \\ 
 \midrule
 TUAB & 23 & 256 & 10 & 409,455 \\
 TUEV & 23 & 256 & 5 & 112,491 \\
 SEED-V & 62 & 1000 & 1 & 148,694 \\
 CHB-MIT & 16 & 256 & 10 & 26,483 \\
 Sleep-EDF & 2 & 100 & 30 & 195,479 \\
 PhysioNet & 64 & 160 & 4 & 18,540 \\
    \bottomrule[1.5pt]
    \end{tabular}}}
\end{table}

 \subsubsection{\method-SP}
To maximize task specification, \method-SP allocates a separate LoRA module, comprising both the down-projection and up-projection matrices, for each task. This approach ensures that the gradients from each task remain entirely distinct and do not interfere with one another. 

Formally, for any linear layer $f$ with weight matrix $W_0 \in \R^{m \times n}$ and bias $b_0$, we define a set of low-rank decomposition matrices $\Delta \mW= \{ \Delta W_i = B_i A_i | B_i \in \R^{m \times r}, A_i \in \R^{r \times n}, i=1, 2, \dots, P\}$ where $r$ is the rank and $P$ is the total number of tasks. When the model performs the $p$-th task, the corresponding adapter is combined with the layer, so the original linear operation is transformed into
 \begin{align*}
     f(x) &= (W_0 + B_p A_p)x + b_0
 \end{align*}
 \subsubsection{\method-RT}
 Inspired by MoE architectures, \method-RT employs the same set of LoRA modules across all tasks, utilizing a router network to dynamically allocate weights to the experts according to the inputs at each layer. This approach promotes the model's ability to identify similarities across various tasks and utilize shared knowledge effectively.

 Formally, for any linear layer $f$ with weight matrix $W_0 \in \R^{m \times n}$ and bias $b_0$, the experts are defined as $\Delta \mW= \{ \Delta W_i = B_i A_i | B_i \in \R^{m \times r}, A_i \in \R^{r \times n}, i=1, 2, \dots, S\}$ where $r$ is the rank and $S$ is the total number of experts. When the model performs any task, a weighted sum of the experts' outputs is combined with the layer, so the original linear operation is transformed into
 \begin{align*}
     f(x) &= (W_0 + \sum_{i=1}^S \omega_i B_i A_i)x + b_0,
 \end{align*}
 where $\omega_i$ denotes the weight of $i$-th expert determined by the router.
 \subsubsection{\method-DC}
 MTEEG-DC decomposes a LoRA module into a common down-projection matrix shared by all the tasks and multiple task-specific up-projection matrices. This approach promotes the dual objectives of global knowledge reuse and task-specific knowledge disentanglement.

 Formally, for any linear layer $f$ with weight matrix $W_0 \in \R^{m \times n}$ and bias $b_0$, the down-projection matrix is denoted as $A$ and the task-specific up-projection matrices are denoted as $\{ B_i \in \R^{m \times r} | i=1, 2, \dots, P\}$ where $r$ is the rank and $P$ is the total number of tasks. When the model performs the $p$-th task, the input is first multiplied by the down-projection matrix $A$, followed by the corresponding up-projection matrix $B_p$. Thus, the original linear operation is transformed into
 \begin{align*}
     f(x) = (W_0 + B_p A)x + b_0
 \end{align*}
 
 We apply the aforementioned transformations to all linear layers in the transformer encoder, including the linear projections of query, key, value and output matrices, as well as the fully connected feed-forward network that follows the attention layers. 
 
 Throughout the fine-tuning stage, all the pre-trained weights are kept frozen and only the low-rank adapters are trainable. In this way, the gradients from different tasks are distinctly separated or confined in different ways, thereby alleviating the heterogeneous conflict issue.

\section{Experiments}

\begin{table*}[th!]\footnotesize
\centering
\caption{Downstream performance of different methods}
\label{tab:results1}
\resizebox{\textwidth}{!}{
\setlength{\tabcolsep}{3mm}{
\renewcommand{\arraystretch}{1.2}
\begin{tabular}{L{2.5cm}|C{1.5cm}C{2.1cm}C{2.2cm}C{2cm}C{2.1cm}C{2.2cm}C{2cm}}
\toprule[1.5pt]
\multirow{2}*{\textbf{Methods}} & \multirow{2}*{\textbf{\# Trainable}} & \multicolumn{3}{c}{\textbf{TUAB}} & \multicolumn{3}{c}{\textbf{TUEV}} \cr
\cmidrule(l){3-5} \cmidrule(l){6-8}
   & \textbf{Params} & Balanced Acc. $\uparrow$ & AUC-PR $\uparrow$ & AUROC $\uparrow$& Balanced Acc. $\uparrow$  & Cohen's Kappa $\uparrow$ & Weighted F1 $\uparrow$  \\
\midrule
\multicolumn{8}{c}{Single-task methods}\\
\midrule
SPaRCNet & 0.79M & 0.7896 & 0.8414 & 0.8676 & 0.4161 & 0.4233 & 0.7024 \\
ContraWR & 1.6M & 0.7746 & 0.8421 & 0.8456 & 0.4384 & 0.3912 & 0.6893 \\
CNN-Transformer & 3.2M & 0.7777 & 0.8433 & 0.8461 & 0.4087 & 0.3815 & 0.6854 \\
FFCL & 2.4M & 0.7848 & 0.8448 & 0.8569 & 0.3979 & 0.3732 & 0.6783 \\
ST-Transformer & 3.5M & 0.7966 & 0.8521 & 0.8707 & 0.3984 & 0.3765 & 0.6823 \\
BIOT & 3.2M & 0.7959 & 0.8792 & 0.8815 & 0.5281 & 0.5273 & 0.7492 \\
LaBraM (Full) & 5.8M & \textbf{0.8126} & 0.8911 & \textbf{0.8843} & \textbf{0.6436} & \textbf{0.6254} & \textbf{0.8172} \\
LaBraM (LoRA) & 0.3M & 0.8105 & \textbf{0.8944} & 0.8824 & 0.6431 & 0.6111 & 0.8103 \\
\midrule
\multicolumn{8}{c}{Multi-task methods}\\
\midrule
HPS & 6.0M & 0.8052 & 0.8740 & 0.8759 & 0.6093 & 0.6097 & 0.8109 \\
 \rowcolor{gray!10}
\method-SP & 1.8M & 0.8096 & 0.8775 & 0.8784 & 0.6438 & 0.6281 & 0.8184 \\ 
 \rowcolor{gray!10}
\method-RT & 1.8M & 0.7964 & 0.8651 & 0.8698 & 0.5574 & 0.5852 & 0.7981 \\
 \rowcolor{gray!10}
\method-DC  & 1.1M & \textbf{0.8118} & \textbf{0.8841}& \textbf{0.8846}&  \textbf{0.6521} & \textbf{0.6398} & \textbf{0.8209}\\
\end{tabular}}}

\resizebox{\textwidth}{!}{
\setlength{\tabcolsep}{3mm}{
\renewcommand{\arraystretch}{1.2}
\begin{tabular}{L{2.5cm}|C{1.5cm}C{2.1cm}C{2.2cm}C{2cm}C{2.1cm}C{2.2cm}C{2cm}}
\toprule[1.5pt]
\multirow{2}*{\textbf{Methods}} & \multirow{2}*{\textbf{\# Trainable}} & \multicolumn{3}{c}{\textbf{SEED-V}} & \multicolumn{3}{c}{\textbf{CHB-MIT}} \cr
\cmidrule(l){3-5} \cmidrule(l){6-8}
   & \textbf{Params} & Balanced Acc. $\uparrow$ & Cohen's Kappa $\uparrow$ & Weighted F1 $\uparrow$& Balanced Acc. $\uparrow$  & AUC-PR $\uparrow$ & AUROC $\uparrow$  \\
\midrule
\multicolumn{8}{c}{Single-task methods}\\
\midrule
SPaRCNet & 0.79M & 0.2865 & 0.1115 & 0.2966 & 0.8417 & 0.9364 & \textbf{0.9151} \\
ContraWR & 1.6M & 0.3681 & 0.2099 & 0.3682 & 0.8034 & 0.9057 & 0.8671 \\
CNN-Transformer & 3.2M & 0.3036 & 0.1367 & 0.2813 & 0.7861 & 0.9032 & 0.8701 \\
FFCL & 2.4M & 0.3714 & 0.2152 & 0.3750 & 0.8106 & 0.9225 & 0.8918 \\
ST-Transformer & 3.5M & 0.2828 & 0.1182 & 0.2740 & 0.8229 & 0.9165 & 0.8942 \\
BIOT & 3.2M & 0.3831 & 0.2238 & 0.3831 & \textbf{0.8439} & \textbf{0.9367} & 0.9026 \\
LaBraM (Full) & 5.8M & \textbf{0.4097} & \textbf{0.2616} & \textbf{0.4119} & 0.8229 & 0.9260 & 0.8989 \\
LaBraM (LoRA) & 0.3M & 0.4056 & 0.2564 & 0.4107 & 0.8244 & 0.9329 & 0.9105 \\
\midrule
\multicolumn{8}{c}{Multi-task methods}\\
\midrule
HPS & 6.0M & 0.4057 & 0.2614 & 0.4108 & 0.7524 & 0.9223 & 0.8914 \\
 \rowcolor{gray!10}
\method-SP & 1.8M & 0.4112 & \textbf{0.2677} & 0.4173 & 0.8586 & 0.9742 & 0.9656 \\
 \rowcolor{gray!10}
\method-RT & 1.8M & 0.4089 & 0.2627 & 0.4153 & 0.7637 & 0.9054 & 0.8925 \\
 \rowcolor{gray!10}
\method-DC  & 1.1M & \textbf{0.4113} & 0.2651 & \textbf{0.4182} & \textbf{0.8657} & \textbf{0.9831} & \textbf{0.9763}\\
\end{tabular}}}

\resizebox{\textwidth}{!}{
\setlength{\tabcolsep}{3mm}{
\renewcommand{\arraystretch}{1.2}
\begin{tabular}{L{2.5cm}|C{1.5cm}C{2.1cm}C{2.2cm}C{2cm}C{2.1cm}C{2.2cm}C{2cm}}
\toprule[1.5pt]
\multirow{2}*{\textbf{Methods}} & \multirow{2}*{\textbf{\# Trainable}} & \multicolumn{3}{c}{\textbf{Sleep-EDF}} & \multicolumn{3}{c}{\textbf{PhysioNet}} \cr
\cmidrule(l){3-5} \cmidrule(l){6-8}
   & \textbf{Params} & Balanced Acc. $\uparrow$ & Cohen's Kappa $\uparrow$ & Weighted F1 $\uparrow$& Balanced Acc. $\uparrow$  & Cohen's Kappa $\uparrow$ & Weighted F1 $\uparrow$  \\
\midrule
\multicolumn{8}{c}{Single-task methods}\\
\midrule
SPaRCNet & 0.79M & 0.7066 & 0.6378 & 0.7538 & \textbf{0.5088} & \textbf{0.4355} & \textbf{0.6253} \\
ContraWR & 1.6M & \textbf{0.7148} & 0.6785 & 0.7837 & 0.3855 & 0.2673 & 0.4888 \\
CNN-Transformer & 3.2M & 0.7095 & \textbf{0.6874} & \textbf{0.7869} & 0.3967 & 0.2986 & 0.5324 \\
FFCL & 2.4M & 0.7143 & 0.6633 & 0.7739 & 0.3868 & 0.2532 & 0.5202 \\
ST-Transformer & 3.5M & 0.6993 & 0.6630 & 0.7690 & 0.4440 & 0.3301 & 0.5433 \\
BIOT & 3.2M & 0.7006 & 0.6740 & 0.7799 & 0.3346 & 0.1642 & 0.3262 \\
LaBraM (Full) & 5.8M & 0.7003 & 0.6742 & 0.7789 & 0.5072 & 0.4303 & 0.6110 \\
LaBraM (LoRA) & 0.3M & 0.6918 & 0.6689 & 0.7759 & 0.5022 & 0.4289 & 0.6086 \\
\midrule
\multicolumn{8}{c}{Multi-task methods}\\
\midrule
HPS & 6.0M & 0.6628 & 0.6411 & 0.7647 & 0.4571 & 0.3677 & 0.5679 \\
 \rowcolor{gray!10}
\method-SP & 1.8M & 0.6847 & 0.6574 & 0.7720 & 0.5087 & 0.4376 & 0.6117 \\
 \rowcolor{gray!10}
\method-RT & 1.8M & 0.6753 & 0.6594 & 0.7678 & 0.4567 & 0.3654 & 0.5660 \\
 \rowcolor{gray!10}
\method-DC  & 1.1M & \textbf{0.7024} & \textbf{0.6754} & \textbf{0.7828} &  \textbf{0.5125} & \textbf{0.4402} & \textbf{0.6143} \\
\bottomrule[1.5pt] 
\end{tabular}}}

\vspace{-0.2cm}
\end{table*}

\begin{figure*}[t]
    \centering
    \includegraphics[width=\linewidth]{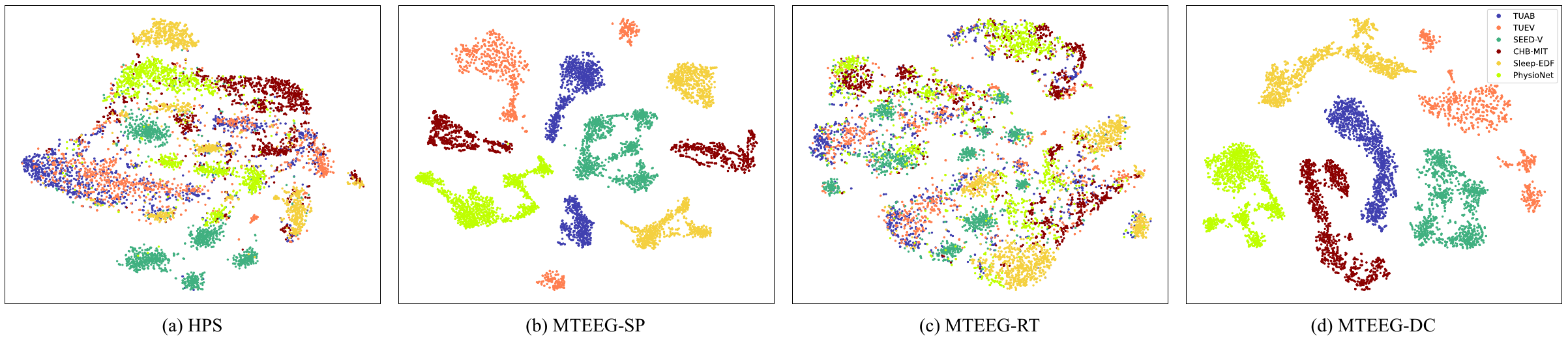}
    \vspace{-0.3cm}
    \caption{Feature distribution of HPS, \method-SP, \method-RT and \method-DC, visualized by t-SNE on the six downstream datasets. The features are extracted from the final layer preceding the classification heads in each model. Results demonstrate that the superior methods in Table \ref{tab:results1} produce features that are more discriminative across different tasks.}
    \vspace{-0.2cm}
    \label{fig:features}
\end{figure*}

\subsection{Downstream Datasets}
After pre-training, we fine-tune and evaluate our \method jointly on the following six datasets, the statistics of which are summarized in Table \ref{tab:downstream}.

\textbf{TUAB} (abnormal detection) \cite{obeid2016temple}: A corpus of EEGs that have been annotated as normal or abnormal.

\textbf{TUEV} (event type classification) \cite{obeid2016temple}: A subset of TUEG that contains annotations of EEG segments as one of six classes: (1) spike and sharp wave (SPSW), (2) generalized periodic epileptiform discharges (GPED), (3) periodic lateralized epileptiform discharges (PLED), (4) eye movement (EYEM), (5) artifact (ARTF) and (6) background (BCKG).

\textbf{SEED-V} (emotion recognition) \cite{liu2021comparing}: An emotion EEG dataset collected while 16 subjects watched video clips corresponding to five emotion categories (happy, sad, neutral, disgust, and fear).

\textbf{CHB-MIT} (seizure detection) \cite{shoeb2009application}: A database from Children’s Hospital Boston consisting of EEG recordings from 22 pediatric subjects with intractable seizures. Signals are sampled with 23 bipolar channels and we select the 16 standard montages in the experiments. Since the dataset is highly imbalanced (about 0.3\% positive ratio), we segment the seizure regions with a 1-second stride to generate overlapping samples. In addition, we follow common practices \cite{lee2024resnet,chung2024single} to randomly select 10\% of the negative samples during training.

\textbf{Sleep-EDF} (sleep stage classification) \cite{goldberger2000physiobank}: A database containing 197 whole-night PolySomnoGraphic sleep recordings, among which we use the 153 recordings from the study of age effects in healthy subjects (SC) in the experiments. Samples are manually annotated as one of the eight classes (W, N1, N2, N3, N4, REM, MOVEMENT, UNKNOWN). Following previous works \cite{supratak2017deepsleepnet,supratak2020tinysleepnet}, we exclude movement artifacts at the beginning and the end of each sleep data that was labeled as MOVEMENT or UNKNOWN, as they do not belong to the ﬁve sleep stages. In addition, we merge the N3 and N4 stages into a single stage N3 to stick to the AASM manual \cite{berry2012aasm}.

\textbf{PhysioNet} (motor imagery classification) \cite{goldberger2000physiobank}: A dataset containing EEG recordings from 109 participants, with trials that belong to 5 classes: left hand, right hand, both hands, both feet, as well as rest. Following previous works \cite{barmpas2023improving,zoumpourlis2024motor}, we discard data from 6 participants (S088, S090, S092, S100, S104, S106) that have inconsistent sampling frequencies or trial lengths.

\subsection{Experimental Setup}
\subsubsection{Data Preprocessing}
Following \cite{jiang2024large}, we first filter the EEG signals within the range of 0.1 Hz to 75 Hz to eliminate low-frequency noise. A 50/60 Hz notch filter is subsequently employed to eliminate power-line interference. After that, all EEG signals are resampled to a frequency of 200 Hz. The typical range of EEG values is between -0.1 mV and 0.1 mV, which we normalize by setting the unit to 0.1 mV to ensure the values predominantly fall between -1 and 1. The same preprocessing pipeline is applied to both the pre-training and downstream datasets.

\subsubsection{Data Split}
For TUAB and TUEV, the training and test sets are provided by the original creator of the dataset. We adhere to BIOT \cite{yang2023biot} and LaBraM to partition the training set into training and validation subsets at a ratio of 80\% and 20\%, respectively. 

For SEED-V, we divide the 15 trials of each session into three groups of five, then consolidate each group from all sessions to create the training, validation, and test sets. 

For CHB-MIT, there are a total of 23 cases collected from 22 subjects. Following BIOT, we use cases 1 to 19 for training, cases 20 and 21 for validation, and cases 22 and 23 for testing. 

For Sleep-EDF and PhysioNet, we partition the recordings by order into training, validation and test sets at a ratio of 64\%, 16\% and 20\%, respectively.

\subsubsection{Training}
In the pre-training of LaBraM, we use the default hyperparameters specified in the original paper, with the exception of the number of temporal embeddings, which we increase from 16 to 64 to accommodate input samples exceeding 16 seconds in duration. The pre-training data comprises nine public datasets, with a total duration of approximately 2000 hours. In the fine-tuning stage, we train the models using binary cross-entropy loss for binary classification tasks and cross-entropy loss for multi-class classification tasks. Due to the significantly larger data volume of TUAB compared to other datasets, which leads to early convergence and overfitting, we randomly sample 10\% of the data points in TUAB for each training epoch to balance the optimization. All the experiments are conducted on Linux servers equipped with NVIDIA A100 GPUs and Python 3.10.14 + PyTorch 2.2.2 + CUDA 12.1 environment. The optimal models are trained on the training set, selected from the validation set, and finally evaluated on the test set. We report the average values on three different random seeds to obtain comparable results.

\subsubsection{Baselines}
For single-task baselines, we consider both self-supervised and supervised methods. Self-supervised baselines include LaBraM and BIOT \cite{yang2023biot}. Supervised baselines include SPaRCNet \cite{jing2023development}, ContraWR \cite{yang2021self}, CNN-Transformer \cite{peh2022transformer}, FFCL \cite{li2022motor} and ST-Transformer \cite{song2021transformer}. LaBraM and BIOT are publicly accessible in their official repositories, with the supervised methods implemented by BIOT. We use the default hyperparameters for fair comparison. 

Given that multi-task learning in EEG processing is underexplored and there is currently no public method for comparison, we incorporate a pre-trained LaBraM as the backbone network within a hard parameter sharing (HPS) \cite{long2017learning,lu2017fully} framework to set up the multi-task baseline. In HPS, different tasks share the same expert (backbone network), except for the classification heads. The implementation is based on LibMTL \cite{lin2022libmtl}.

\subsubsection{Metrics}
Following \cite{jiang2024large} and \cite{yang2023biot}, we use Balanced Accuracy, AUC-PR and AUROC for binary classification tasks and Balanced Accuracy, Cohen's Kappa and Weighted F1 for multi-class classification tasks. The implementation of all the metrics are based on PyHealth \cite{pyhealth2023yang}.

\subsection{Comparison with Prior Works}

The main results are summarized in Table \ref{tab:results1}. Firstly, there exists a notable performance gap between HPS and LaBraM across all tasks and metrics, despite their architectural similarities. This suggests that, similar to other data types, EEG signals from diverse sources can also confuse the model due to conflicting optimization directions, resulting in substantial performance degradation. Secondly, our proposed \method-SP and \method-DC significantly outperform HPS across all tasks, demonstrating the efficacy of gradient separation with task-specific low-rank modules. Furthermore, \method-DC performs better than the state-of-the-art single-task methods on four out of six tasks. This indicates that the decomposition of LoRA into task-agnostic and task-specific matrices helps the model benefit from both task interaction and specification, yielding better representations for downstream performance. On the other hand, the performance of \method-RT is subpar compared to the other two variants, which contradicts the effectiveness of MoE-based approaches in other domains. This may stem from the router being implemented with basic linear layers, which may be inadequate for differentiating the intricate intrinsic properties of highly noisy EEG signals. Thirdly, \method has the advantage of being lightweight. The three variants have a maximum of 1.8M trainable parameters during fine-tuning. The efficiency of this lightweight design would be beneficial in practical applications, particularly when resources are limited.

\begin{table}[t]
    \centering
    \caption{Ablation Study on the LoRA Rank $r$}
    \label{ablation:r}
    \resizebox{0.48\textwidth}{!}{
    \setlength{\tabcolsep}{0.7mm}{
    \hspace*{-0.3cm}
    \begin{tabular}{l|C{1.5cm}C{1.5cm}C{1.5cm}C{1.5cm}C{1.5cm}C{1.5cm}}
    \toprule[1.5pt]
  & {TUAB} & {TUEV}  & {SEED-V} & {CHB-MIT} & {SleepEDF} & {PhysioNet} \\ 
 \midrule
\multicolumn{7}{c}{\method-SP}\\
\midrule
 r=4 &0.8065 &0.6242 & 0.4021&0.8242 & 0.6797 & 0.4937\\
 r=8 &\textbf{0.8096} &\textbf{0.6438} & \textbf{0.4112}&\textbf{0.8586}& \textbf{0.6847} & 0.4956\\
 r=16 &0.8080 &0.6217 &0.4093&0.8468& 0.6780 & 0.4856\\
 r=32 &0.7985 &0.5839 & 0.4042&0.8266 & 0.6816 & \textbf{0.4985}\\
 \midrule
 \multicolumn{7}{c}{\method-DC}\\
\midrule
 r=4 &0.8064 & 0.6303 & 0.4036 & 0.8371 & 0.6961 & 0.4963\\
r=8 &\textbf{0.8118} & \textbf{0.6521} & \textbf{0.4113} & \textbf{0.8657} & \textbf{0.7024} & \textbf{0.5125}\\
r=16 & 0.8076 & 0.6280 & 0.4074 & 0.8548 & 0.6944 & 0.5118\\
r=32 & 0.8019 & 0.6107 & 0.4069 & 0.8322 & 0.7003 & 0.4998\\
    \bottomrule[1.5pt]
    \end{tabular}}}
    \vspace{-0.2cm}
\end{table}

\begin{table}[t]
    \centering
    \caption{Ablation Study on Adapter Locations}
    \label{ablation:loc}
    \resizebox{0.48\textwidth}{!}{
    \setlength{\tabcolsep}{0.7mm}{
    \hspace*{-0.3cm}
    \begin{tabular}{l|C{1.5cm}C{1.5cm}C{1.5cm}C{1.5cm}C{1.5cm}C{1.5cm}}
    \toprule[1.5pt]
  & {TUAB} & {TUEV}  & {SEED-V} & {CHB-MIT} & {SleepEDF} & {PhysioNet} \\ 
 \midrule
\multicolumn{7}{c}{\method-SP}\\
\midrule
 MHSA & 0.8049 & 0.6038 & 0.3989 & 0.8454& 0.6833 & 0.4878\\
 FFN & 0.8075 & 0.6131 & 0.3947 & 0.8239 & 0.6815 & 0.4952\\
 Both &\textbf{0.8096} &\textbf{0.6438} & \textbf{0.4112}&\textbf{0.8586}& \textbf{0.6847} & \textbf{0.4956}\\
 \midrule
 \multicolumn{7}{c}{\method-DC}\\
\midrule
MHSA & 0.7989 & 0.6280 & 0.4062 & 0.8400 & 0.6952 & 0.5110\\
FFN & 0.8086 & 0.6214 & 0.4026 & 0.8351 & 0.6984 & 0.5080\\
Both &\textbf{0.8118} & \textbf{0.6521} & \textbf{0.4113} & \textbf{0.8657} & \textbf{0.7024} & \textbf{0.5125}\\
    \bottomrule[1.5pt]
    \end{tabular}}}
    \vspace{-0.2cm}
\end{table}

\subsection{Feature Visualization}
The primary goal of \method is to alleviate potential conflicts between different tasks, so we assert that the resulting representational space should show some task-specific patterns. To validate this, we randomly select 1280 samples from the test set of each task (dataset) and extract the corresponding features from the final layers preceding the classification heads in each model. These features are subsequently visualized with t-SNE. As shown in Figure \ref{fig:features}, \method-SP and \method-DC produce more discriminative features across different tasks than \method-RT, which supports their stronger performance in Table  \ref{tab:results1}. Furthermore, the boundaries between tasks produced by \method-DC are more notable than those produced by \method-SP. This indicates that task interaction, which is promoted by the incorporation of a task-agnostic down-projection matrix, is beneficial for the model's ability to distinguish between tasks and reduce task interference.

\subsection{Ablation Studies}

We perform ablation studies on two factors that may have impact on the model's performance: the LoRA rank $r$ and the locations where LoRA modules are applied. In the ablation studies, balanced accuracy is used as the primary metric for comparison.

\subsubsection{Impact of adapter rank $r$}
We assign different values to $r$, ranging from 4 to 32 to examine its impact on the model's downstream performance. As illustrated in Table \ref{ablation:r}, \method-DC consistently achieves its maximum performance at $r=8$ across all datasets, whereas \method-SP reaches peak performance at $r=32$ on PhysioNet and $r=8$ on the remaining datasets. This indicates that a higher rank does not necessarily yield better performance, likely due to over-fitting induced by an excess of parameters. Therefore, we select $r=8$ as the default configuration in our experiments.

\subsubsection{Impact of adapter locations}
The selection of locations for applying low-rank adapters is known to significantly influence the model's performance \cite{hu2021lora}. Thus, we evaluate three different configurations of adapter locations in the transformer encoder: (1) only in multi-head self-attention modules (MHSA), (2) only in the feed-forward networks (FFN) that follow MHSA, (3) in both MHSA and FFN. As shown in Table \ref{ablation:loc}, the adaptations of both MHSA and FFN are crucial, as the elimination of either leads to a significant decline in performance.

\section{Conclusion}
This paper presents \method, an innovative multi-task EEG analysis framework. Utilizing a powerful pre-trained model, \method incorporates LoRA modules to disentangle the parameter spaces for different tasks, thereby alleviating the conflicts stemming from the heterogeneity of EEG signals. We propose three variants of \method that combines the LoRA modules in different ways to take account of different degrees of task specification and interaction, and validate their effectiveness on six publicly available datasets. Experiments show that \method can simultaneously manage abnormal detection, event type classification, emotion recognition, seizure detection, sleep stage classification and motor imagery classification, outperforming state-of-the-art single-task methods on most tasks and metrics. The versatility of \method demonstrate the significant potential of multi-task EEG analysis and promote the advancement of general-purpose brain-computer interfaces in the future.


\bibliographystyle{IEEEtran}
\bibliography{citations}

\end{document}